\documentclass[conference]{IEEEtran}
\IEEEoverridecommandlockouts

\usepackage{cite}
\usepackage{amsmath,amssymb,amsfonts}
\usepackage{amsthm}
\usepackage{algorithmic}
\usepackage{graphicx}
\usepackage{textcomp}
\usepackage{xcolor}

\usepackage[linesnumbered,ruled,vlined]{algorithm2e}
\usepackage{fancyhdr}
\usepackage{graphicx}
\usepackage{float}
\usepackage{subfig}
\usepackage{multirow}
\usepackage{booktabs}
\usepackage{array}
\theoremstyle{definition}
\newtheorem{definition}{Definition}

\newtheorem{proofOfTheorem}{Proof of Theorem}
\newtheorem{theorem}{Theorem}

\def\BibTeX{{\rm B\kern-.05em{\sc i\kern-.025em b}\kern-.08em
    T\kern-.1667em\lower.7ex\hbox{E}\kern-.125emX}}
\begin{document}

\title{Federated Hypergraph Learning with Local Differential Privacy: Toward Privacy-Aware Hypergraph Structure Completion}

\author{\IEEEauthorblockN{Linfeng Luo, Zhiqi Guo, Fengxiao Tang*, Zihao Qiu, Ming Zhao}
\IEEEauthorblockA{\textit{School of Computer Science and Engineering} \\
\textit{Central South University}\\
\{luolinfeng,guozq,tangfengxiao,qiuzh,meanzhao\}@csu.edu.cn}\\
\thanks{\textsuperscript{*}Corresponding author}
}

\maketitle

\begin{abstract}

The rapid growth of graph-structured data necessitates partitioning and distributed storage across decentralized systems, driving the emergence of federated graph learning to collaboratively train Graph Neural Networks (GNNs) without compromising privacy. However, current methods exhibit limited performance when handling hypergraphs, which inherently represent complex high-order relationships beyond pairwise connections. Partitioning hypergraph structures across federated subsystems amplifies structural complexity, hindering high-order information mining and compromising local information integrity.
To bridge the gap between hypergraph learning and federated systems, we develop FedHGL, a first-of-its-kind framework for federated hypergraph learning on disjoint and privacy-constrained hypergraph partitions. Beyond collaboratively training a comprehensive hypergraph neural network across multiple clients, FedHGL introduces a pre-propagation hyperedge completion mechanism to preserve high-order structural integrity within each client. This procedure leverages the federated central server to perform cross-client hypergraph convolution without exposing internal topological information, effectively mitigating the high-order information loss induced by subgraph partitioning. Furthermore, by incorporating two kinds of local differential privacy (LDP) mechanisms, we provide formal privacy guarantees for this process, ensuring that sensitive node features remain protected against inference attacks from potentially malicious servers or clients.
Experimental results on seven real-world datasets confirm the effectiveness of our approach and demonstrate its performance advantages over traditional federated graph learning methods.
\end{abstract}


\section{Introduction}
Graph neural networks (GNNs) are widely used in fields such as recommendation systems and network analysis due to their ability to capture complex relationships within graph-structured data \cite{INFOCOM-HTNet}. However, the rapidly growing volumes and complexity for graph-structured data necessitate distributed storage, while strict data protection regulations hinder information sharing across systems \cite{FedGNN-survey}. Federated learning is designed to address the challenges of training neural networks in distributed systems, enabling participants to collaboratively build a shared model while preserving privacy and security by avoiding direct data sharing \cite{convergence}. To facilitate structured data mining across clients, researchers have proposed federated graph learning, which enables collaborative training of a mutual GNN model while preserving privacy by avoiding direct data sharing between data owners \cite{arXiv-fedgnn, NIPS-FedSage, NIPS-FedGCN, TPDS-FedCog}. 

While federated learning on simple graphs has been extensively studied, the scenario changes significantly when it comes to hypergraphs. As a type of complex graph structure, Hypergraphs enable the connection of multiple nodes through a single hyperedge. Compared to simple graphs, hypergraphs offer advantages by capturing higher-order relationships that reflect the multi-dimensional interconnectivity present in various real-world data structures. Recently, Hypergraph Neural Networks have gained prominence as tools for mining features and patterns on complex graph-structured data in data-rich environments \cite{AAAI-HGNN,NIPS-HyperGCN,AAAI-DHCN, arXiv-hnhn}.

\begin{figure}[!t]
	\centering
	\subfloat[]{\includegraphics[width=0.45\columnwidth]{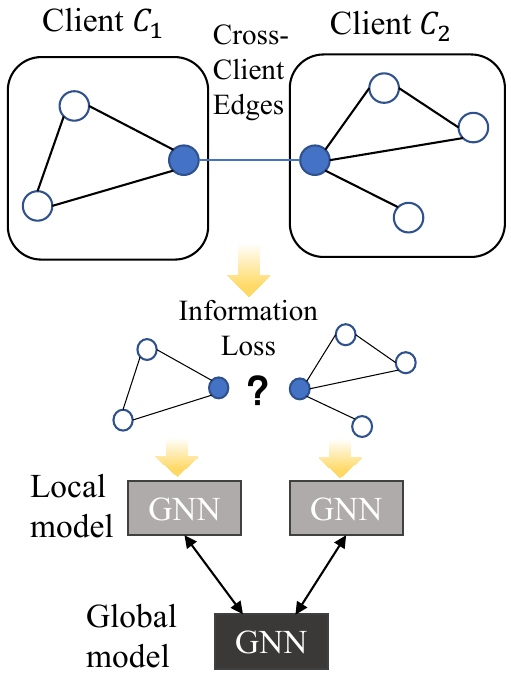}%
		\label{fig:simple}}
	\hfil
	\subfloat[]{\includegraphics[width=0.45\columnwidth]{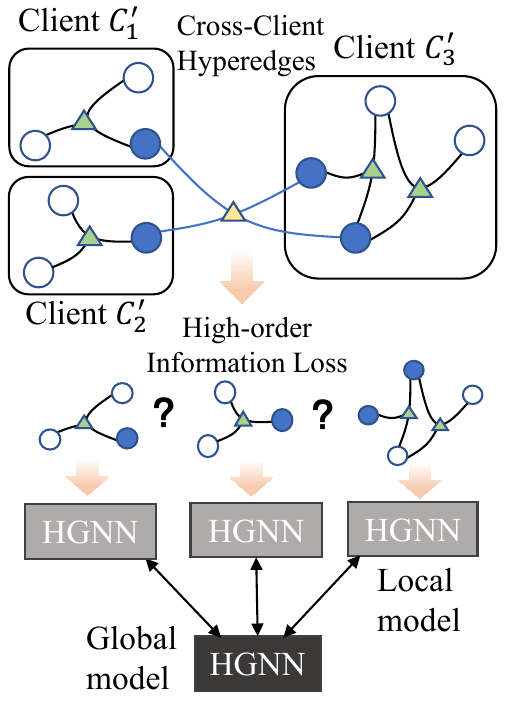}%
		\label{fig:hyper}}
	\caption{Cross-client information loss occurs when a simple graph or a hypergraph is stored in a distributed manner.}
	\label{fig:challenge}
\end{figure}

In this work, we focus on the scenario of federated graph learning where each client possesses a subgraph of either a simple graph or a hypergraph. Similar to simple graphs, real-world hypergraph data is often distributed across decentralized data sources, calls for federated hypergraph learning methods. For instance, family relationships among patients in different hospitals, or customers holding accounts across different banks, can both naturally form hypergraph structures distributed across different data owners. However, traditional federated graph learning methods are not applicable to hypergraph due to the following reasons:

\begin{itemize}

\item Traditional GNNs do not support hypergraph structures;

\item Distributed data storage causes loss of high-order information between multiple clients;

\item Communications between clients lack privacy protection.

\end{itemize}

Obviously, federated graph learning cannot handle hypergraph mining tasks on clients using standard GNNs. Another critical issue lies in the loss of cross-client information, which arises because no single client can leverage edge information across clients, as illustrated in Fig. \ref{fig:challenge}. However, mechanisms designed to address this information loss on simple graphs cannot be directly applied to hypergraphs. These methods generally fall into two categories: One is FedSage+ \cite{NIPS-FedSage} and other generative methods, where the core idea is to mend the missing neighbors across subgraphs. However, the presence of hyperedges complicates the mending process beyond pairwise connections, making it difficult to effectively share the necessary gradient information for training a missing-node generator. Another approach is FedGCN \cite{NIPS-FedGCN}, which uses the federated central server as a trusted third party to help clients compute the propagated features of border nodes. However, outsourcing the computation of local border nodes to a third party may expose sensitive topological structures or node attributes to malicious clients or servers, making it unsuitable for applications with strict topology privacy requirements.

In this work, we propose FedHGL, a novel federated hypergraph learning algorithm designed to address the above challenges. In FedHGL, independent HGNN models~\cite{AAAI-HGNN} are deployed on each client as the graph mining model, and the global model parameters are then updated and synchronized to the clients using the FedAVG \cite{FedAVG} algorithm. Furthermore, we introduce a pre-propagation hyperedge completion (HC) operation to mitigate the loss of high-order information across clients. This pre-training process is not intended to shift the computational mission from clients to the federated central server, but rather to enable the server to compute the inherently cross-client shared component—namely, the feature representations of hyperedges—through a one-time computation prior to training. Therefore, we advance and decompose the hypergraph neural network propagation process: The central server first acts as a cross-client hypergraph convolution executor to compute the features of corresponding cross-client hyperedges; then, the aggregated hyperedge features are returned to the clients as cross-client information supplements. Finally, local differential privacy (LDP) is employed to protect client data during the multi-party communication and aggregation process, preventing leakage to malicious servers or clients.

Our main contributions can be summarized as follows:

\begin{itemize}

\item We are the first to present the comprehensive solution for federated hypergraph learning across isolated and privacy-sensitive subgraph. 

\item We introduce the HC process, a pre-propagation mechanism that enables cross-client hypergraph convolution in federated settings.
     
\item By incorporating two types of local differential privacy, we ensure the privacy of border node features during the HC process against malicious clients or servers.

\item We tested FedHGL on real-world hypergraph datasets and compared it with state-of-the-art federated subgraph learning methods on simple graph datasets, demonstrating its optimal performance.

\end{itemize}

\section{Related Work}

The field of hypergraph learning has evolved considerably in recent years \cite{HGNN-survey}. The groundwork for hypergraph learning was laid in \cite{NIPS-hypergraph}. Building upon this, \cite{AAAI-HGNN} extended spectral convolution to hypergraphs by proposing Hypergraph Neural Networks (HGNN). Subsequently, their work was extended to the domain of dynamic graphs, leading to the development of Dynamic Hypergraph Neural Networks (DHGNN) \cite{DHGNN}.  Another contribution is HyperGCN by \cite{NIPS-HyperGCN}, which simplifies the learning process by approximating hyperedges with pairwise edges. Incorporating additional structural information, \cite{arXiv-hnhn} proposed Hypergraph Neural Networks with Line Expansion (HNHN), which enhances performance through a more detailed representation of hyperedges. In \cite{arXiv-HyperSage}, authors introduced HyperSAGE, an inductive framework that generalizes representation learning on hypergraphs using a two-level neural message passing strategy. The dual-channel approach to hypergraph convolution is developed by \cite{AAAI-DHCN} with Dual Channel Hypergraph Convolutional Networks (DHCN), leveraging two separate channels for node and hyperedge updates. 

Differential privacy (DP) was initially developed to protect individual privacy when publishing aggregated or statistical data \cite{DP}. This approach adds random noise to data query results, ensuring that changes to individual information in the dataset do not significantly alter the distribution of the output. However, in distributed data collection scenarios, Local Differential Privacy (LDP) is more suitable because it applies noise directly on the user's device, eliminating the need for a trusted central authority. In practical applications, Google \cite{Rappor} and Samsung \cite{Harmony} employ Local Differential Privacy (LDP) to gather anonymized user data, thereby enhancing user's privacy without sacrificing service quality. Currently, existing researchs have employed DP or LDP mechanisms to ensure the security of model parameter sharing in federated learning \cite{dp-fed, ldp-fed,ldp-fed2}. 

Federated subgraph learning has gained significant attention for enabling collaborative learning across distributed subgraphs. FedGNN, introduced by \cite{arXiv-fedgnn}, is the first federated subgraph learning framework designed to preserve user privacy in recommendation systems. In another effort, \cite{ICLR-fedgraphnn} developed FedGraphNN, a comprehensive benchmark system for federated learning with GNNs. \cite{NIPS-FedSage} proposed FedSage+, a federated subgraph learning method with a missing neighbor generator to address incomplete neighbor information in federated settings. \cite{TPDS-FedGraph} introduced FedGraph, which enhances graph learning capabilities through intelligent sampling and cross-client convolution operations while preserving privacy. In \cite{NIPS-FedGCN}, FedGCN has been proposed and reduces communication overhead and improves convergence rates in federated training of graph convolutional networks by using homomorphic encryption and differential privacy techniques. \cite{TPDS-FedCog} presented FedCog, a federated learning framework for coupled graphs that efficiently manages distributed graph data and improves node classification performance. 

Despite the advancements, these methods are unsuitable for federated hypergraph learning because the higher-order relationships in hypergraphs necessitate different modeling techniques and propagation rules. Furthermore, these methods do not provide a comprehensive solution when addressing the issue of cross-client high-order information loss.
\section{Federated Hypergraph Learning}

In this section, we first introduce the notations used in federated hypergraph learning, then define the problem of semi-supervised node classification. Finally, we present the base version of our FedHGL framework.

\subsection{Preliminary}

\begin{definition}[Hypergraph]
Let $G=(V,E,\mathbf{W},\mathbf{X})$ denotes a hypergraph, in which $V$ is a set containing $|V|$ vertices and $E$ is a set containing $|E|$ hyperedges. $\mathbf{W}\in\mathbb{R}^{|E|\times |E|}$ is the diagonal weight matrix of hyperedges where $w(e_{j})=W_{jj}$ represents the weight of hyperedge $e_{j}\in E$. $\mathbf{X}\in\mathbb{R}^{|V|\times P}$ is the feature matrix where vector $\mathbf{x}_{v}\in\mathbb{R}^{P}$ denotes $P$-dimensional features of the vertex $v\in V$.
\end{definition}

We define the connection between node $v_i$ and edge $e_j$ as:
\begin{eqnarray}\label{Eq:hve}
	h(v_i,e_j)=\left\{
	\begin{aligned}
		1& , & if\,v_i \in e_j,\\
		0& , & if\,v_i \notin e_j
	\end{aligned}
	\right.
\end{eqnarray}

In hypergraph $G$, the connection between nodes and edges can be represented by an incidence matrix $\mathbf{H}\in\mathbb{R}^{|V|\times |E|}$, where $H_{i,j}=h(v_i,e_j)$. On this basis, the degree of $v_{i}$ is defined as $d_{v_i}=\sum_{e_j\in E}w(e_j)h(v_i,e_j)$, and the diagonal matrix $\mathbf{D}_V\in\mathbb{R}^{|V|\times |V|}$ represents degrees of each vertex. The degree of $e_{i}$ is define as $d_{e_i}=\sum_{v_j\in V}h(v_j,e_i)$, and the degrees of each hyperedge form the diagonal matrix $\mathbf{D}_E\in\mathbb{R}^{|E|\times |E|}$.

\begin{definition}[$\epsilon$-Local Differential Privacy]\label{def:LDP}
A perturbation algorithm $\mathcal{P}$ satisfies $\epsilon$-local differential privacy, where $\epsilon\geq 0$, if and only if for any pair of input attribute $A,A^{\prime}\in Domain(\mathcal{P})$ and any possible output $A^*\in Range(\mathcal{P})$, we have
\begin{equation}\label{eq:ldp}
\frac{Pr[\mathcal{P}(A)=A^*]}{Pr[\mathcal{P}(A^{\prime})=A^*]}\leq e^{\epsilon}.
\end{equation}
\end{definition}

\subsection{Problem Setup}

We take semi-supervised node classification as the task for federated hypergraph learning. In this context, we assume there is a central server $\mathcal{S}$ and $K$ clients $C=\{c_k \mid k \in \mathbb{Z}^+ , k < K\}$ involved, and each client $c_k$ has access to a subgraph $G_k=(V_k,E_k,\mathbf{W}_k,\mathbf{X}_k)$ of the hypergraph $G$. The subgraph information stored by clients is exclusive, which implies that for any two different subgraphs $G_i$ and $G_j$, not only do $V_{i}\cap V_{j}=\emptyset$ and $E_i\cap E_j=\emptyset$, but also $\forall{e_i\in E_i,e_j \in E_j}, e_i\cap e_j=\emptyset$. We ignore the case where there is any node in $G$ that does not belong to any client, i.e., $V=\bigcup_{k=1}^{K}V_{k}$. The hyperedges that include nodes from different clients are termed cross-client hyperedges, represented by $E^{*} = E - \bigcup_{k=1}^{K}E_k$. The subset of $E^{*}$ accessible to client $c_k$ is denoted by $E_k^*$, and the border nodes of client $c_k$ that connect to $E_k^*$ are indicated by $V_k^*$.

In the semi-supervised node classification task, only the nodes in a subset have one-hot labels that represent the types of the nodes in the hypergraph $G$. For client $c_k$, the nodes used for training are denoted as $\mathcal{V}_k \subseteq V_k$, and their labels are represented by $\mathbf{Y}_k$. The task of semi-supervised node classification involves using $G_k$ and $\mathbf{Y}_k$ to train a local node classification model $F(\mathbf{\Theta}_k)$, inferring the remaining unknown labels on the subgraph. $\mathbf{\Theta}_k$ represents the learnable parameter matrix of the hypergraph learning model. Our main purpose in the context of federated hypergraph learning is to develop a global node classifier, denoted as $F(\mathbf{\Theta})$. Specifically, the optimization objective of federated hypergraph learning is to minimize the global empirical loss $\mathcal{R}$:
\begin{eqnarray}\label{Eq:mr}
\min_{\mathbf{\Theta}}\mathcal{R}(\mathbf{\Theta}):=\min_{\mathbf{\Theta}}\sum_{k=1}^K\frac{\vert{\mathcal{V}_k}\vert}{\vert{\mathcal{V}}\vert}\mathcal{R}_k(\mathbf{\Theta}).
\end{eqnarray}

In Eq. \ref{Eq:mr}, $\vert{\mathcal{V}_k}\vert$ represents the number of nodes in $\mathcal{V}_k$, and $\vert{\mathcal{V}}\vert = \sum_{k=1}^K{\vert{\mathcal{V}_k}\vert}$. $\mathcal{R}_k$ represents the local empirical loss for client $c_k$. We define $\mathcal{R}_{k}$ as:

\begin{eqnarray}\label{Eq:ri}
\begin{split}
\mathcal{R}_{k}(\mathbf{\mathbf\Theta}):=\mathcal{L}(F(G_k,E_k^*;\mathbf{\Theta}),\mathbf{Y}_k)\\
:=\frac{1}{\vert{\mathcal{V}_k}\vert}\sum_{v\in \mathcal{V}_k}\mathcal{L}(F_v(G_k,E_k^*;\mathbf{\Theta}),y_v),
\end{split}
\end{eqnarray}
where $\mathcal{L}$ represents the loss function, and $F_v$ denotes the output of classifier $F$ on node $v$.

\subsection{Basic FedHGL}

\newcommand{\INPUT}{\item[\textbf{Input:}]}
\newcommand{\OUTPUT}{\item[\textbf{Output:}]}
\newcommand{\COMMENTABOVE}[1]{\STATE $\triangleright$ #1}

\begin{algorithm}[t]
\caption{Federated Hypergraph Learning (FedHGL)}
\label{alg:FedHGL}
\DontPrintSemicolon  
\SetKwInput{Input}{Input}
\SetKwFor{For}{for}{do}{end}
\SetKwIF{If}{ElseIf}{Else}{if}{then}{else if}{else}{end}

\Input{Server $\mathcal{S}$, clients $C$, subgraphs of hypergraph $G$ $\{G_k\}$, labels $\{Y_k\}$, border hyperedges $E^*$, learning rate $\eta$, perturbation algorithm $\mathcal{P}$.}
\tcp*[l]{On client $c_k$}
\For{$c_k \in C$} {

    \eIf{$is\_HC=True$}{
        $\mathbf{X}^{(N)}_k\gets HC(\mathcal{S},C,\{G_k\},E^*,\mathcal{P})$
    }
    {
        $\tilde{E}^*_k\gets\text{Trim}(E_k^*,V_k^*)$\;
    }

}

\tcp*[l]{On server $\mathcal{S}$}

initiate $\mathbf{\Theta}^{0}=\{\mathbf{\Theta}^{(n)}\}$\;
    \For{$t=0$ \KwTo $T-1$}{
        Broadcast $\mathbf{\Theta}^{t}$ to $C$\;
        \tcp*[l]{On client $c_k$}
        \For{$c_k \in C$} {
        
            $\mathbf{\Theta}_k \gets \mathbf{\Theta}^t$\;
            \For{$i = 1$ \KwTo $I$}{
                
                $\mathbf{\Theta}_k \gets \mathbf{\Theta}_k - \eta \nabla \mathcal{L}(F(\mathbf{\Theta}_k;G_k,\tilde{E}^*_k),\mathbf{Y}_k)$\;
            }
            $\mathbf{\Theta}_k^{t+1} \gets \mathbf{\Theta}_k$\;
            Send $\mathbf{\Theta}_k^{t+1}$ to server $\mathcal{S}$\;
        }
        $\mathbf{\Theta}^{t+1} \gets \sum_{k=1}^{K}\frac{\vert{\mathcal{V}_k}\vert}{\vert{\mathcal{V}}\vert}\mathbf{\Theta}_k^{t+1}$\;
    }

\end{algorithm}

To perform semi-supervised node classification on subgraphs of a hypergraph, we propose a federated hypergraph learning algorithm, FedHGL, as detailed in Algorithm~\ref{alg:FedHGL}. At each client, we employ HGNN as the hypergraph mining model, which, as previously mentioned, performs hypergraph convolution in two stages: hyperedge feature gathering and node feature aggregation. A hyperedge convolutional layer in HGNN can be formulated by:

\begin{eqnarray}\label{Eq:xl1}
\mathbf{X}^{(n+1)}=\sigma(\mathbf{D}_V^{-\frac{1}{2}}\mathbf{H}\mathbf{W}\mathbf{D}_E^{-1}\mathbf{H}^{T}\mathbf{D}_V^{-\frac{1}{2}}\mathbf{X}^{(n)}\mathbf{\Theta}^{(n)}),
\end{eqnarray}
where $\sigma$ represents the non-linear activation function, and $\mathbf{\Theta}^{(n)}$ is the convolution filter parameter matrix at the $n$-th layer. In FedHGL, our classifier contains $N$ layers of HGNN. The HGNN model employs spectral convolution on hypergraphs and a node-edge-node transformation method to aggregate node features via hyperedges, thereby effectively enhancing data representation and feature extraction. We denote $\mathbf{x}_v^{(n)}$ as the representation of node $v\in V_k$ caculated by the $n$-th HGNN layer in client $c_k$,and graph convolution can be viewed as feature propagation between nodes:

\begin{eqnarray}\label{Eq:xv}
\mathbf{x}_v^{(n+1)}=\sigma(\sum_{e\in E_k\cup \tilde{E}^*_k}{\sum_{u\in V_k}\frac{w(e)h(v,e)h(u,e)}{d_e\sqrt{d_v d_u}}\mathbf{x}_u^{(n)}}\mathbf{\Theta}_{k}^{(n)}).
\end{eqnarray}

\begin{algorithm}[t]
\caption{HC: Hyperedge Completion}
\label{alg:HC}
\DontPrintSemicolon  

\SetKwInOut{Input}{Input}
\SetKwInOut{Output}{Output}

\Input{Server $\mathcal{S}$, clients $C$, subgraphs of a hypergraph $\{G_k\}$, border hyperedges $E^*$, perturbation algorithm $\mathcal{P}$}
\Output{Node embeddings $\{\mathbf{X}^{(N)}_k$\}}

\For{$n = 0$ \KwTo $N-1$}{
    \tcp{On client $c_k$}
    \For{$c_k \in C$}{
        \For{$e^* \in E_k^*$}{
            $\delta^{(n)}(e^*, V_k^*) \gets Eq.~\ref{Eq:e} $\;
            \eIf{$n = 0$ and $h(e^*,V_k)=1$}{
                $\delta^{(n)}_{c_k,e^*} \gets \mathcal{P}(\delta^{(n)}(e^*, V_k^*))$
            }
            {
                $\delta^{(n)}_{c_k,e^*} \gets \delta^{(n)}(e^*, V_k^*)$
            }
            Send $\delta^{(n)}_{c_k,e^*}$ to $\mathcal{S}$\;
        }
    }
    \tcp{On server $\mathcal{S}$}
    \For{$e^* \in E^*$}{
        Receive $\{\delta^{(n)}_{c,e^*}\mid c\in {C}_{e^*}\}$ from $\mathcal{C}_{e^*}$
        $\delta^{(n)}(e^*, V^*) \gets \sum_{c\in C_{e^*}}\delta^{(n)}_{c,e^*} $\;
        Send $\delta^{(n)}(e^*, V^*)$ to $\mathcal{C}_{e^*}$\;
    }
    \tcp{On client $c_k$}
    \For{$c_k \in C$}{
        \For{$v \in V_k$}{
            $\mathbf{x}_{v}^{(n+1)} \gets Eq.~\ref{Eq:xv2}$\;
        }
    }
}
\Return{$\{\mathbf{X}^{(N)}_k\}$}
\end{algorithm}

Note that client $c_k$ cannot directly access the features of other clients' nodes connected by $E_k^*$. Instead of dropping the cross-client hyperedges, we trim the cross-client hyperedges $E_k^*$ to $\tilde{E}^*_k$ by removing nodes not included in $V_k$. Thus, $c_k$ preserves the structure of cross-client hyperedges locally.

In every training round, client $c_k$ updates the parameter matrix of the $n$-th HGNN layer locally for $I$ iterations by $\mathbf{\Theta}_k^{(n)}\gets \mathbf{\Theta}_k^{(n)}-\eta\nabla \mathcal{R}_k(\mathbf{\Theta}_k)$, where $\eta$ is the learning rate. We choose cross-entropy as the loss function. Then, $c_k$ uploads $\mathbf{\Theta}_k^{(n)}$ to the central server which uses the FedAvg algorithm to aggregate the global parameter matrix by $\mathbf{\Theta}^{(n)}\leftarrow\sum_{k=0}^K\frac{\vert{\mathcal{V}_k}\vert}{\vert{\mathcal{V}}\vert}\mathbf{\Theta}_k^{(n)}$. The updated global parameters are then sent down to each client to update their local models.

\begin{figure*}[t]
\centering
\includegraphics[width=0.95\textwidth]{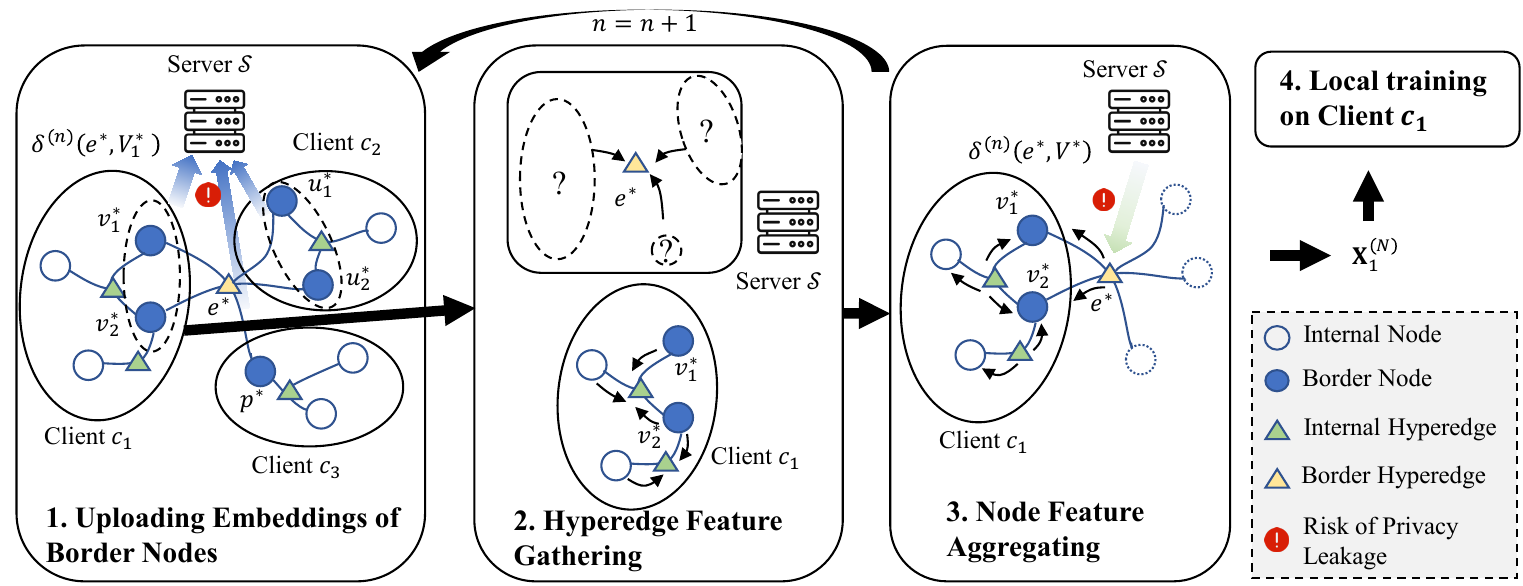} 
  \caption{\textbf{Pre-propagation hyperedge completion on the subgraph:} In HC, the feature gathering step for border hyperedges is transferred to the central server without requiring knowledge of the clients' internal structures.}
\label{fig:alg}
\end{figure*}

\section{Hyperedge Completion and Security Concerns}

In the basic version of FedHGL, individual clients are unable to access the features of nodes from other clients connected through cross-client hyperedges, resulting in incomplete representations of subgraphs and high-order information loss. In this section, we introduce HC process and two LDP mechanisms, enhancing FedHGL by supplementing cross-client information for the clients while avoiding privacy leaks due to extra communications. 


\subsection{Hyperedge Completion on Subgraphs}

In conventional federated graph learning methods, clients typically compute message passing for border nodes using cross-client information—either by fetching required features from other clients or by offloading the computation to a third party. The former risks exposing sensitive data to untrusted clients, while the latter poses a risk of exposing internal client topology. This motivates our central idea: instead of offloading any client-specific computation, we delegate only the part of message passing involving cross-client structures to the federated central server.

Based on this insight, we perform two kinds of decomposition of the original HGNN propagation process. In the first decomposition, we split the computation of node embeddings into two steps: edge feature gathering and node feature aggregation. Consequently, in the $n$-th HGNN layer, hyperedges gather the embeddings of connected nodes by:

\begin{eqnarray}\label{Eq:e}
\delta^{(n)}(e,V)=\sum_{u\in V}\frac{h(u,e)}{\sqrt{d_u}}\mathbf{x}_{u}^{(n)},
\end{eqnarray}
the embeddings of nodes are then calculated by aggregating the embeddings of their related hyperedges in the $(n+1)$-th layer HGNN:

\begin{eqnarray}\label{Eq:phi}
\phi^{(n+1)}(v,E,V)=\sum_{e\in E}\frac{w(e)h(v,e)}{d_e\sqrt{d_v}}\delta^{(n)}(e, V).
\end{eqnarray}

Then, we expect no information loss when nodes aggregate features from hyperedges; therefore, the second decomposition is performed where the embeddings of nodes are computed by:


\begin{eqnarray}\label{Eq:xv2}
\mathbf{x}_{v}^{(n+1)}=\phi^{(n+1)}(v, E_k,V_k)+\phi^{(n+1)}(v, {E}_k^*,V^*),
\end{eqnarray}
where $V^*=\bigcup_{k=1}^{K}{V_k^*}$ denotes all border nodes across the clients. Eq. \ref{Eq:xv2} reveals that for the non-border nodes, the computation of $\phi^{(n)}(v, {E}_k^*, V^*)$ is unnecessary. Conversely, the border nodes must aggregate features not only from their local client but also from adjacent border nodes across various clients linked via cross-client hyperedges to compute $\delta^{(n)}$.

After the decompositions of the original HGNN propagation process, we can formally represent the HC process. As presented in Algorithm~\ref{alg:HC} and Fig.~\ref{fig:alg}, we introduce $N$ rounds of pre-propagation hyperedge completion in FedHGL, with each round corresponding to one HGNN layer computed prior to formal training. For the $n$-th round, clients compute the embeddings $\delta^{(n)}(e^*, V_k^*)$ for each cross-client hyperedge $e^* \in E_k^*$, and upload them to the central server. We assume that each client is aware of the existence of cross-client neighbors linked to its border nodes via cross-client hyperedges, but does not know their feature values. Thus, the server can gathers the embeddings $\delta^{(n)}(e^*, V^*)$ based on each cross-client hyperedge's identifier without knowing which nodes are border nodes, and distributes them back to the clients connected to those hyperedges. After that, the clients use the received embeddings to locally compute the features of their border nodes. The aggregated embeddings of border nodes will be used in the $(n+1)$-th round pre-propagation. 


\begin{table}[!t]

\centering
\caption{Comparison of additional communication overhead caused by cross-client information supplementation.}

\begin{tabular}{@{}cc@{}}

\toprule
Model         & Communication Overhead \\ \midrule
FedSage+      &  $O(TNK\vert V^*\vert f)$             \\
FedCog        &  $O(NK\vert V^*\vert f)$      \\
FedGCN(a-hop) &  $O(N\vert V^*\vert^a f)$                \\
FedHGL        &  $O(NK\vert E^*\vert f)$           \\ \bottomrule
\end{tabular}

\label{tab:overhead}
\end{table}

\subsection{Overhead Analysis}

Based on the node embeddings calculated by the HC process, client $ c_k $ can train the classifier $ F(\mathbf{\Theta}_k) $. Notably, we remove the activation functions between multiple HGNN layers, which allows the propagation of node embeddings to be computed just once across multiple training iterations. The output of the classifier $F$ for node $v$ can be denoted as:

\begin{eqnarray}\label{Eq:fv}
F_v( G_k, {E}_k^*;\mathbf{\Theta}_k)=\sigma\left(\mathbf{x}_v^{(N)}\mathbf{\Theta}_{k}\right),
\end{eqnarray}
where the learnable parameter matrix $\mathbf{\Theta}_{i}$ is the product of the parameters across multiple HGNN layers. 

In Table \ref{tab:overhead}, we compare the additional communication overhead introduced by several federated graph learning algorithms with cross-client information supplementation schemes and our HC operation. Here, $f$ represents the dimension of node features, $\vert V^*\vert$ denotes the number of border nodes, $\vert E^*\vert$ represents the number of boundary hyperedges, and $T$ refers to the number of communication rounds for parameter update in federated learning. FedSage+ incurs the highest overhead as it requires supplementation of information regarding border nodes during formal training. In FedHGL, $K$ serves to conceal internal client structures: by uploading locally aggregated features for each cross-client hyperedge, the central server can perform aggregation without knowing which internal nodes are connected to the hyperedge.

As for the additional computational overhead on the server, FedGCN and FedHGL incur complexities of $O(|V^*|)$ and $O(|E^*|)$, respectively, focusing on the computation of border node features and cross-client hyperedge features. When complex high-order relationships exist across clients, FedHGL demonstrates a certain advantage.


\subsection{LDP based Hyperedge Completion}

Although HC prevents the exposure of clients’ internal topologies to a malicious server, the risk of privacy leakage still exists. As illustrated in Fig.~\ref{fig:alg}, during the HC operation, clients must share information related to border node features with the server and other clients, which introduces the potential for internal node features to be exposed. In FedHGL, the objective of node feature privacy protection is to ensure that a client’s features are not disclosed to any malicious server or client, assuming the possibility of adversarial collusion. To evaluate the privacy leakage risks introduced by the HC operation, we identify a necessary condition under which the features of border nodes could be exposed to a malicious server $\mathcal{\dot{S}}$ or a set of malicious clients $\dot{C}$:

\begin{theorem}
\label{condition1}
The necessary condition for a victim client $ c_k $ to leak node features to a malicious central server $\mathcal{\dot{S}}$ or a set of malicious clients $\dot{C}$ in HC is that: there exists $e^*\in E^*$ such that $h(e^*,V_k)=1$, where $h(e,V)=\sum_{v\in V}{h(e,v)}$.  
\end{theorem}

\begin{proofOfTheorem}
\label{proof}
In the first round of HC, if Theorem~\ref{condition1} is not satisfied, it implies that the content $\delta^{(0)}(e^*, V_k^*)$, uploaded by the victim client $c_k$  contains the sum of the features from more than one node in $V_k^*$. Clearly, $\mathcal{\dot{S}}$ is unable to extract the original features $\mathbf{x}_{v^*}^{(0)}$ of any single node $v^* \in V_k^*$ from $\delta^{(0)}(e^*, V_k^*)$. Assuming that all clients except $c_k$ are malicious and can share information among themselves, $\dot{C}$ can thus compute the features uploaded by $c_k$ from the cross-client hyperedge features distributed by the server by:
\begin{equation}\label{Eq:deltacross}
\delta^{(0)}(e^*, V_k^*)=\delta^{(0)}(e^*, V^*)-\delta^{(0)}(e^*, V^*\setminus V_k^*).
\end{equation}
Likewise, $\dot{C}$ cannot calculate the original features $\mathbf{x}_{v^*}^{(0)}$ of any individual node $v^* \in V_k^*$ from $\delta^{(0)}(e^*, V_k^*)$.

From the second round onward, clients upload intermediate HGNN embeddings instead of raw features. As these embeddings are non-invertible and abstracted, reconstructing original features becomes infeasible, thus mitigating privacy risks.

\end{proofOfTheorem}

Theorem~\ref{condition1} reveals that the HC operation poses a risk of privacy leakage for the node features only when a cross-client hyperedge connects to a single node $ v^* $ within the victim client $ c_k $. This is intuitive, as when a cross-client hyperedge connects multiple nodes within the victim client, the uploaded representation corresponds to the aggregated features of these border nodes, making it infeasible to infer the feature of any individual node. Therefore, we only need to introduce the privacy protection mechanism when the necessary condition is met, in order to achieve the goal of safeguarding node privacy. LDP operates by adding noise to each user's data prior to upload, making it challenging to infer any individual's data through aggregate analysis. We denote the node feature that needs to be uploaded as $ \delta_{c_k, e^*}^{(n)} $. When the condition in Theorem~\ref{condition1} holds, it implies the following:
\begin{equation}\label{Eq:dnew}
\delta_{c_k,e^*}^{(0)}=\mathcal{P}(\delta^{(0)}(e^*, V_k^*)) = \mathcal{P}\left(\frac{\mathbf{x}_{v^*}^{(0)}}{\sqrt{d_{v^*}}}\right).
\end{equation}
Assuming each node has $f$ features, we represent each feature as $A_1, A_2, \dots, A_f$. By introduces the perturbation algorithm $ \mathcal{P} $ to individually perturb each attribute $d_{v^*}^{-\frac{1}{2}}\mathbf{x}_{v^*}^{(0)}[A_j]$ before it is uploaded, inferring the original node features of a victim client from the uploaded perturbed data becomes difficult for malicious servers or clients, even when the necessary condition in Theorem~\ref{condition1} is satisfied.

This work primarily focuses on two types of node features: binary attributes and numeric attributes. We use randomized response (RR) \cite{RR} and the Laplace mechanism \cite{DP} as perturbation schemes for these two scenarios.

\subsubsection{Randomized Response}
The original RR mechanism reports the true value from a user with a probability of $ p $; with a probability of $ 1 - p $, it reports one of two binary values, each with equal likelihood. Thus, the probability that the node reports the true value is $ (1+p)/2 $, and the probability of reporting the false value is $ (1-p)/2 $.

However, in original RR setting, $ \delta_{c_k,e^*}^{(0)}[A_j]$ is not an unbiased estimate of the regularized node feature $d_{v^*}^{-\frac{1}{2}} \mathbf{x}_{v^*}^{(0)}[A_j]$. We assume that the original node features take values of $0$ or $1$; then, the regularized feature $d_{v^*}^{-\frac{1}{2}}\mathbf{x}_{v^*}^{(0)}[A_j]$ takes values of 0 or $d_{v^*}^{-\frac{1}{2}}$. The expected value of the attribute reported by the node to the central server in the original RR setting is $pd_{v^*}^{-\frac{1}{2}}\mathbf{x}_{v^*}^{(0)}[A_j]+(1-p)d_{v^*}^{-\frac{1}{2}}/2$. In the aggregation process of edge hyperedge features, we aim for the features provided by clients to more accurately reflect the general characteristics of the boundary nodes. Therefore, we need to adjust the reported attributes to unbiasedness. In the HC process, if the condition in Theorem~\ref{condition1} is satisfied, $ \delta^{(0)}(e^*, V_k^*)[A_j] $ will be sampled from the distribution:

\begin{equation}\label{Eq:delta1}
Pr[\delta_{c_k,e^*}^{(0)}[A_j]=x] = 
\begin{cases} 
    \frac{1+p}{2}, & \text{if } x = \frac{p+2\mathbf{x}_{v^*}^{(0)}[A_j]-1}{2p\sqrt{d_{v^*}}} \\ 
    \frac{1-p}{2}, & \text{if } x = \frac{p-2\mathbf{x}_{v^*}^{(0)}[A_j]+1}{2p\sqrt{d_{v^*}}}
\end{cases}.
\end{equation}
To satisfy $\epsilon$-LDP as defined in Definition \ref{def:LDP}, the probability condition $\frac{1+p}{1-p} \leq e^\epsilon $ must be met; therefore, we set $p = \frac{e^\epsilon - 1}{e^\epsilon + 1}$.

\subsubsection{Laplace Mechanism}
To implement LDP using the Laplace mechanism, we perturb each attribute of the features uploaded by the node as follows:
\begin{equation}\label{Eq:delta2}
\delta_{c_k,e^*}^{(0)}[A_j]=d_{v^*}^{-\frac{1}{2}}\mathbf{x}_{v^*}^{(0)}[A_j]+Lap(\frac{s}{\epsilon}),
\end{equation}
where $Lap(\frac{s}{\epsilon})$ denotes a random noise following a Laplace distribution with a scale parameter $\frac{s}{\epsilon}$, characterized by the following probability density function:
\begin{equation}
pdf(x)=\frac{\epsilon}{2s}exp\left(-\frac{\epsilon |x|}{s}\right),
\end{equation}
and the sensitivity $s$ depends on the range of the attribute values:
\begin{equation}\label{Eq:s}
s=d_{v^*}^{-\frac{1}{2}}\max_{v^*,u^*\in V^*_i}|\mathbf{x}_{v^*}^{(0)}[A_j]-\mathbf{x}_{u^*}^{(0)}[A_j]|,
\end{equation}
and each perturbed attribute is an unbiased estimate of the original attribute, as the expected value of the added Laplace noise is zero.

\section{Experiments}

In this section, we validate the effectiveness of FedHGL through ablation experiments, compare it with state-of-the-art federated subgraph learning methods, and explore the impact of different LDP mechanisms on performance to balance privacy and efficiency.

\subsection{Performance on Hypergraph}
\subsubsection{Datasets}

For the semi-supervised node classification task on subgraphs of hypergraphs, we use four hypergraph datasets provided by the DHG (DeepHypergraph) library, as shown in Table \ref{tab:hdata}. These datasets include CoraCA from \cite{NIPS-HyperGCN} and DBLP4k from \cite{dblp}, both citation network datasets; a movie network dataset IMDB4k from \cite{WWW-imdb}; and a newspaper network dataset 20Newsgroups from \cite{ICLR-allset}. In DBLP4k, the hyperedges are constructed by the co-paper correlation and co-term correlation, and in IMDB4k, the hyperedges are constructed by the co-director correlation and the co-actor correlation. We refer to \cite{NIPS-FedGCN} to partition data based on labels by using the Dirichlet distribution and set $\beta=10000$ to simulate the i.i.d. setting, where number of clients is set to $K=3, 6, 9$.

\subsubsection{Experimental Settings}

We compare our FedHGL algorithm (with and without HC) against non-federated training methods on the hypergraph datasets: Local HGNN where there is no communication between clients, local HGNN with HC operation, and Global HGNN where a single client uses all information of the graph. Additionally, we compared FedHGL with two other hypergraph models training in federated manner: federated HyperGCN \cite{NIPS-HyperGCN} and federated HNHN \cite{arXiv-hnhn}, which evidently lack a cross-client information supplementation mechanism. We set the number of HGNN layers to $N=2$ and hidden features to $16$ with drop rate $p=0.5$. Following the settings of GCN \cite{ICLR-GCN}, in the transductive node classification task, only a portion of the nodes have labels, and only a small number of samples are used for training. Therefore, we set the validation-testing ratio on each client to $20\%\slash 40\%$ and adjust the training ratio based on the number of nodes and label classes, as shown in Table \ref{tab:hdata}. We use Adam Optimization to minimize our cross-entropy loss function with a learning rate of 0.01 (Adam optimizer).

\begin{table}[t]

\centering
\small
\caption{Statistics of the hypergraph datasets, where $\vert{E^*}\vert$ represents the number of border hyperedges, and $\vert{v^*}\vert$ represents the number of unsafe border node.}

\begin{tabular}{@{}c|c|c|c|c|c@{}}

\toprule
\multicolumn{2}{c|}{Dataset}     &  CoraCA  & DBLP4k &IMDB4k & 20News \\ \midrule
\multicolumn{2}{c|}{Nodes}     &  2708   & 4057     & 4278   & 16342  \\
\multicolumn{2}{c|}{Hyperedges}  &  1072 & 22051     & 7338   & 100    \\
\multicolumn{2}{c|}{Classes}    &   7   &  4        & 3      & 4      \\
\multicolumn{2}{c|}{Features}   &  1433  &    334   & 3066   & 1433   \\ 
\multicolumn{2}{c|}{Training Ratio} &  0.1  &    0.06    & 0.06    & 0.01    \\ 
\multicolumn{2}{c|}{Data Type} & Binary & Numeric & Numeric & Numeric \\ \midrule
\multicolumn{1}{c|}{\multirow{2}{*}{$K=3$}}& $\vert{E^*}\vert $ &   820 &   5320    & 2379   & 100    \\ 
\multicolumn{1}{c|}{} & $\vert{v^*}\vert $ &   852 &   2354    &  2243  &   0  \\ \midrule
\multicolumn{1}{c|}{\multirow{2}{*}{$K=6$}}&$\vert{E^*}\vert $ &  894 &  5813      & 2606   & 100    \\ 
\multicolumn{1}{c|}{} & $\vert{v^*}\vert $ &  1319  &   2932    & 3252  &   0  \\ \midrule
\multicolumn{1}{c|}{\multirow{2}{*}{$K=9$}}&$\vert{E^*}\vert$  & 921 &   5967     & 2678   & 100    \\ 
\multicolumn{1}{c|}{} & $\vert{v^*}\vert $ & 1606   &  3197     &  3593  &  0   \\ \bottomrule
\end{tabular}

\label{tab:hdata}
\end{table}

\subsubsection{Results and Discussion}

The experimental results of semi-supervised node classification on four hypergraph datasets are shown in Table \ref{tab:hacc}. Accuracy is shown outside the parentheses; the inside indicates the variance-based confidence interval. Under different datasets and client number settings, our proposed FedHGL achieves optimal node classification accuracy. Before performing HC operation, the basic version of FedHGL outperformed the independently trained local HGNN models on the clients, with an average improvement of 5.9\%. The performance gap between FedHGL and local hypergraph models demonstrates the benefits of joint training across multiple clients. However, compared with other federated learning-trained hypergraph models, the basic FedHGL does not show a comprehensive advantage.

After conducting HC operation, FedHGL is further improved by 6.8\% after introducing HC, reducing the performance drop compared with the global HGNN from an average of 10.3\% to 3.5\%. Meanwhile, compared with two other federated hypergraph models: federated HyperGCN and federated HNHN, FedHGL with HC achieved average improvements of 12.8\% and 14.9\%, respectively. The performance improvement brought by the HC operation indicates the impact of cross-client information loss and the effectiveness of our pre-propagation operation, which can be observed from the comparison between local HGNN models with and without the HC operation. 

\begin{table*}[t]
\caption{Node Classification results on the hypergraph datasets compared with non-federated methods and other federated hypergraph learning models.}
\centering

\begin{tabular*}{\linewidth}{p{1.26cm}<{\centering}p{0.945cm}<{\centering}p{0.945cm}<{\centering}p{0.945cm}<{\centering}p{0.945cm}<{\centering}p{0.945cm}<{\centering}p{0.945cm}<{\centering}p{0.945cm}<{\centering}p{0.945cm}<{\centering}p{0.945cm}<{\centering}p{0.945cm}<{\centering}p{0.945cm}<{\centering}p{0.945cm}<{\centering}}
\toprule
Dataset                                                                                              & \multicolumn{3}{c}{CoraCA}                               & \multicolumn{3}{c}{DBLP4k}                       & \multicolumn{3}{c}{IMDB4k}                      & \multicolumn{3}{c}{20News}                      
                 \\ \midrule
Model                                                                                                & K=3               & K=6               & K=9               & K=3               & K=6               & K=9                                                                                                        & K=3               & K=6               & K=9               & K=3               & K=6               & K=9               \\ \cmidrule(lr){2-4} \cmidrule(lr){5-7}
       \cmidrule(lr){8-10} \cmidrule(lr){11-13}
\multirow{2}{*}{HGNN}                                                                          & 0.5353            & 0.3778            & 0.3155            & 0.7643            & 0.7006            & 0.6470                                                                              & 0.4379            & 0.3758            & 0.3481            & 0.7630            & 0.7245            & 0.7027            \\
                                                                                                     & (0.0234)    & (0.0226)    & (0.0250)    & (0.0229)    & (0.0274)    & (0.0268)                                                                                                      & (0.0181)    & (0.0135)    & (0.0168)    & (0.0120)    & (0.0131)    & (0.0210)    \\
\multirow{2}{*}{\begin{tabular}[c]{@{}c@{}}HGNN\\with HC\end{tabular}} & 0.6369            & 0.5545            & 0.5104            & 0.7810            & 0.7065            & 0.6529         & 0.4882            & 0.4466            & 0.4191            & 0.7646            & 0.7297            & 0.7076            \\                                                                            & (0.0262)    & (0.0297)    & (0.0231)    & (0.0265)    & (0.0351)    & (0.0352)   & (0.0224)    & (0.0201)    & (0.0179)    & (0.0135)    & (0.0123)    & (0.0206)   \\ \midrule
\multirow{2}{*}{\begin{tabular}[c]{@{}c@{}}Fed-HNHN\end{tabular} }                                                                      & $0.5047$ & $0.4015$ & $0.3460$ & $0.7380$ & $0.6947$ & $0.6887$ & $0.4045$ & $0.3686$ & $0.3562$ & $0.7365$ & $0.7225$ & $0.7164$ \\
                                                                                                     & (0.0315)    & (0.0253)    & (0.0203)    & (0.0288)    & (0.0397)    & (0.0345)   & (0.0297)    & (0.0226)    & (0.0231)    & (0.0146)    & (0.0144)    & (0.0154)     \\ 
\multirow{2}{*}{\begin{tabular}[c]{@{}c@{}}Fed-\\HyperGCN\end{tabular}  }                                                                      & $0.5896$ & $0.5044$ & $0.4783$ & $0.6182$ & $0.5694$ & $0.5404$  & $0.4357$ & $0.4337$ & $0.4243$ & $0.6529$ & $0.6048$ & $0.5785$ \\
                                                                                                     & (0.0204)    & (0.0346)    & (0.0236)    & (0.0214)    & (0.0189)    & (0.0188)      & (0.0187)    & (0.0221)    & (0.0177)    & (0.0207)    & (0.0195)    & (0.0317)  \\ \midrule
\multirow{2}{*}{ \begin{tabular}[c]{@{}c@{}}FedHGL\end{tabular}}                                                                       & 0.5831            & 0.4703            & 0.3815            & 0.7734            & 0.7263            & 0.7083        & 0.4625            & 0.4088            & 0.3727            & 0.7832            & 0.7755            & 0.7699           \\
                                                                                                     & (0.0229)    & (0.0322)    & (0.0246)    & (0.0443)    & (0.0620)    & (0.0400)    & (0.0242)    & (0.0195)    & (0.0167)    & (0.0077)    & (0.0102)    & (0.0153)   \\
\multirow{2}{*}{\begin{tabular}[c]{@{}c@{}}FedHGL\\with HC\end{tabular}}                                                                      & $\mathbf{0.6900}$ & $\mathbf{0.6726}$ & $\mathbf{0.6585}$ & $\mathbf{0.7859}$ & $\mathbf{0.7481}$ & $\mathbf{0.7249}$  & $\mathbf{0.5329}$ & $\mathbf{0.5325}$ & $\mathbf{0.5292}$ & $\mathbf{0.7843}$ & $\mathbf{0.7793}$ & $\mathbf{0.7746}$ \\
                                                                                                      & (0.0248)    & (0.0282)    & (0.0291)    & (0.0410)    & (0.0606)    & (0.0407)  & (0.0198)    & (0.0205)    & (0.0200)    & (0.0077)    & (0.0099)    & (0.0138)   \\ \cmidrule(lr){2-4} \cmidrule(lr){5-7} \cmidrule(lr){8-10} \cmidrule(lr){11-13}                                                                                                                                                                                                         
Global       & \multicolumn{3}{c}{0.6953 (0.0216)}                 & \multicolumn{3}{c}{0.8515 ($0.0187$)}       & \multicolumn{3}{c}{0.5413 (0.0205)}                 & \multicolumn{3}{c}{0.7888 (0.0073)}                  

                                                                          \\ \bottomrule
\end{tabular*}
\label{tab:hacc}
\end{table*}

\subsection{Performance on simple graph}

\subsubsection{Datasets}

\begin{table}[h]
\centering
\small
\caption{Statistics of the simple graph datasets, where $\vert{\mathcal{E}^*}\vert$ represents the number of cross-client edges, $\vert{E^*}\vert$ represents the number of border hyperedges and $\vert{v^*}\vert$ represents the number of unsafe border node.}

\begin{tabular}{@{}p{2.5cm}<{\centering}|c|p{1.5cm}<{\centering}|p{1.5cm}<{\centering}|p{1.5cm}<{\centering}@{}}
\toprule
\multicolumn{2}{c|}{Dataset}                 & Cora  & CiteSeer & Facebook \\ \midrule
\multicolumn{2}{c|}{Nodes}                   & 2708  & 3327   & 22470 \\
\multicolumn{2}{c|}{Edges}                   & 10858 & 9464  & 85501  \\
\multicolumn{2}{c|}{Hyperedges}              & 2590  & 2996  & 22407  \\
\multicolumn{2}{c|}{Classes}                 & 7     & 6     & 4  \\
\multicolumn{2}{c|}{Features}                & 1433  & 3703   & 4714  \\
\multicolumn{2}{c|}{Training Ratio}   &  0.1  &    0.1    &  0.008  \\ \midrule
\multicolumn{1}{c|}{\multirow{3}{*}{$K=3$}} &$\vert{\mathcal{E}^*}\vert$  & 3463      &  2986   & 56855    \\
\multicolumn{1}{c|}{}                     &$\vert{E^*}\vert$  &  2351     & 2496   &  20373    \\
\multicolumn{1}{c|}{}                     &$\vert{v^*}\vert$  &   1501    &  2118  &   11457   \\\midrule
\multicolumn{1}{c|}{\multirow{3}{*}{$K=6$}} &$\vert{\mathcal{E}^*}\vert$  &  4623     &  3988  &  74868    \\
\multicolumn{1}{c|}{}                     &$\vert{E^*}\vert$  &   2528    &  2821   &  21811   \\
\multicolumn{1}{c|}{}                     &$\vert{v^*}\vert$  &   2234    &  2780  &   17563   \\\midrule
\multicolumn{1}{c|}{\multirow{3}{*}{$K=9$}} &$\vert{\mathcal{E}^*}\vert$  &  4623     &  3988  &  74868    \\
\multicolumn{1}{c|}{}                     &$\vert{E^*}\vert$  &   2528    &  2821   &  21811   \\
\multicolumn{1}{c|}{}                     &$\vert{v^*}\vert$  &   2439    & 2983   &  19714    \\\midrule
\end{tabular}
\label{tab:sdata}
\end{table}

By utilizing potential higher-order relationships to generate hypergraphs, FedHGL can be applied to simple graphs datasets to compare with state-of-the-art federated subgraph learning methods. we deploy the FedHGL and other algorithms on the citation network datasets Cora and CiteSeer \cite{Cora-Citeseer}, and a social network dataset Facebook \cite{facebook}. The details of these datasets are shown in Table \ref{tab:sdata}. We follow \cite{NIPS-FedGCN} to partition the data by leveraging the Dirichlet distribution, setting $\beta=10000$ to simulate the i.i.d. scenario. The number of clients is configured as $K=3, 6, 9$. The generation of hyperedges is achieved through the nearest 1-hop neighbors method on the simple graphs, as referenced in \cite{TPAMI-hypergraph}. 

\begin{table*}[t]
\centering
\small
\caption{Node Classification results on simple graph datasets compared with SOTA federated subgraph learning methods.}
\begin{tabular*}{0.89\linewidth}{@{}ccccccccccc@{}}
\toprule
Dataset                         & \multicolumn{3}{c}{Cora}                                          & \multicolumn{3}{c}{CiteSeer}                                      & \multicolumn{3}{c}{Facebook}                                      \\ \midrule
Model                           & K=3                             & K=6            & K=9                 & K=3                             & K=6              & K=9               & K=3                & K=6             & K=9                             \\ \cmidrule(l){2-4} \cmidrule(l){5-7} \cmidrule(l){8-10} 
\multirow{2}{*}{FedSage}        & 0.6858                          & 0.6032          &  0.5602                & 0.6299                          & 0.6018        & 0.5989                  & 0.7141                          & 0.6295              & 0.5883           \\
                                & (0.0242)                  & (0.0259)        &     (0.0211)      & (0.0166)                  & (0.0204)       &           ( 0.025)                   & (0.0126)                  & (0.0193)           &    (0.0211)   \\
\multirow{2}{*}{FedGCN (0-hop)}  & 0.7272                   & 0.6474  &  \underline{0.5907} & \underline{0.6568} & \underline{0.6307}   & \underline{0.6173}  & 0.7348                          & 0.6647               &  0.6234         \\
                                & (0.0190)                  & (0.0314)      & (0.0262)             & (0.018)                  & (0.0171)                  & ( 0.0167)                & (0.0182)                  & (0.0175)      & (0.0239)             \\
\multirow{2}{*}{FedHGL}  & \underline{0.7565} & \underline{0.6514}                         & 0.5802                          & 0.5818                         & 0.4793 & 0.4165 & \underline{0.8046} & \underline{0.7562} & \underline{0.7261} \\
                                & (0.0149)                  & (0.0258)       & (0.0208)            & (0.0198)        & (0.0186)           & (0.0209)                  & (0.0084)                  & (0.0119)       & (0.0253)            \\ \cmidrule(l){2-4} \cmidrule(l){5-7} \cmidrule(l){8-10} 
\multirow{2}{*}{FedSage+}       & 0.8120                          & 0.8013           &    0.7910           & 0.6901                          & 0.6887        &  0.6791              & 0.7935                          & 0.7755              & 0.7569           \\
                                & (0.0171)                  & (0.0159)      & (0.0189)              & (0.0189)                  & (0.0178)                  & ( 0.0196)              & (0.0134)                  & (0.0141)         & (0.0172)           \\
\multirow{2}{*}{FedGCN (2-hop)}  & 0.8277                          & 0.8256           & 0.8239              & 0.7073                          & 0.6993              &  0.6985           & 0.8219                          & 0.8151          & 0.8131      \\
                                & (0.0139)                  & (0.0141)         & (0.0160)          & (0.0156)                  & (0.0169)                  & ( 0.0227)                         & (0.0108)                  & (0.0074)            & (0.0117)       \\
\multirow{2}{*}{FedCog}         & 0.8178                          & 0.8186                & 0.8212                          & 0.7034            &    0.7038          & 0.7030                          & 0.8078          &   0.8171 &  0.8117           \\
                                & (0.0199)                  & (0.0177)     & (0.0151)             & (0.0154)                  & (0.0185)                  & ( 0.0192)             & (0.0105)                  & (0.0079)     & (0.0086)             \\
\multirow{2}{*}{FedHGL with HC} & $\mathbf{0.8352}$               & $\mathbf{0.8286}$     & $\mathbf{0.8246}$           & $\mathbf{0.7076}$               & $\mathbf{0.7074}$               & $\mathbf{0.7061}$    & $\mathbf{0.8409}$               & $\mathbf{0.8396}$          & $\mathbf{0.8394}$      \\
                                & (0.0205)                  & (0.0137)        & (0.0153)          & (0.0139) & (0.0212)                                & (0.0192)               & (0.0092) & (0.0065)   & (0.0094)  \\
\bottomrule
\end{tabular*}
\label{tab:sacc}
\end{table*}
\subsubsection{Experimental Settings}

We choose FedSage \cite{NIPS-FedSage}, FedGCN \cite{NIPS-FedGCN} and FedCog \cite{TPDS-FedCog} as baseline methods for our study on simple graph datasets. These subgraph federated learning methods address the issue of cross-client edge information loss in simple graphs. Specifically, FedSage, which employs GraphSage model \cite{NIPS-graphsage} locally, and FedGCN (0-hop), which shares 0-hop neighbor information, both ignore any cross-client information loss between clients, similar to FedHGL without HC. Meanwhile, FedSage+ generates missing nodes for clients through additional training; FedGCN (2-hop) uploads information from two neighbors of the target nodes to the server, where it computes the target node's embeddings. These two methods address cross-client information loss issues similar to FedHGL with HC. FedCog uses the SGC \cite{ICML-SGC} model and achieves federated subgraph learning without information loss through graph decoupling operations. All hypergraph neural network layers in these methods are set to 2, with 16 hidden features and a drop rate of $( p = 0.5 )$, to get the optimal performance. We set the validation$\slash$testing ratio to $20\%\slash 40\%$, and use Adam optimizer to minimize our cross-entropy loss function with a learning rate of 0.01.

\subsubsection{Results and Discussions}

As shown in Table \ref{tab:sacc}, the experimental results of FedHGL for semi-supervised node classification on three simple graph datasets demonstrate its optimal performance. Before incorporating the HC operation, FedHGL did not have an advantage over other methods that ignore cross-client information loss. One potential reason is that we generate hyperedges by selecting the 1-hop neighbors from the simple graph without introducing other high-order information. However, the HC operation supplements missing cross-client information, allowing FedHGL to recover and surpass the performance of the current state-of-the-art federated subgraph learning methods. FedHGL shows better performance compared to other federated subgraph learning methods with cross-client information supplementation, outperforming FedSage+ by 2.8\%, FedGCN (2-hop) by 0.8\%, and FedCog by 1\%. The results show that FedHGL not only addresses hypergraph mining tasks where traditional methods are not applicable, but also provides a superior solution for handling cross-client information loss in federated subgraph learning.

\subsection{Tradeoff on privacy and performance}

\subsubsection{Experimental Settings}
The above discussion demonstrates the superior performance of our FedHGL. However, when a high level of privacy protection is required, perturbing the features of border nodes inevitably leads to a decline in performance. Users of the FedHGL algorithm must adjust the privacy budget based on specific requirements to achieve a balance between algorithm performance and the level of privacy protection. A higher privacy budget implies smaller perturbations, leading to less impact on algorithm performance but a lower level of privacy protection. Conversely, a lower privacy budget introduces greater noise, which compromises performance in exchange for stronger privacy guarantees. 

To evaluate the performance of the LDP mechanisms under varying privacy budgets, we selected four datasets: the binary-type hypergraph dataset CoraCA, the numeric-type hypergraph dataset DBLP, the binary-type simple graph dataset Cora, and the numeric-type simple graph dataset Facebook. In the Cora and CoraCA datasets, each dimension of the node features represents a 0/1 word vector, indicating the absence or presence of a corresponding word in a scientific paper. The number of clients is set to $ K = 6 $.

\subsubsection{Results and Discussions}

\begin{figure}[t]
    \centering
    \subfloat[CoraCA (Binary).]{%
        \includegraphics[width=1.6in]{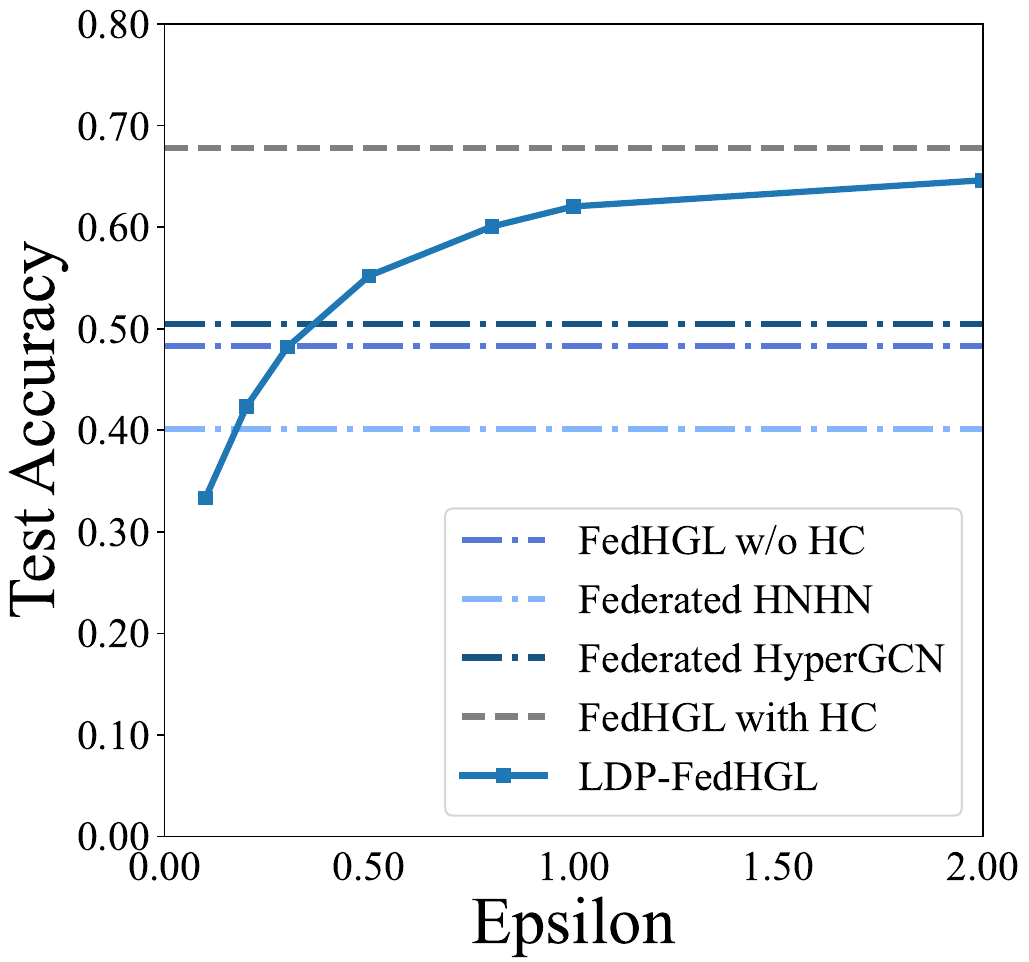}
        \label{fig:CoraCAE}
    }
    \hfil
    \subfloat[DBLP (Numeric).]{%
        \includegraphics[width=1.6in]{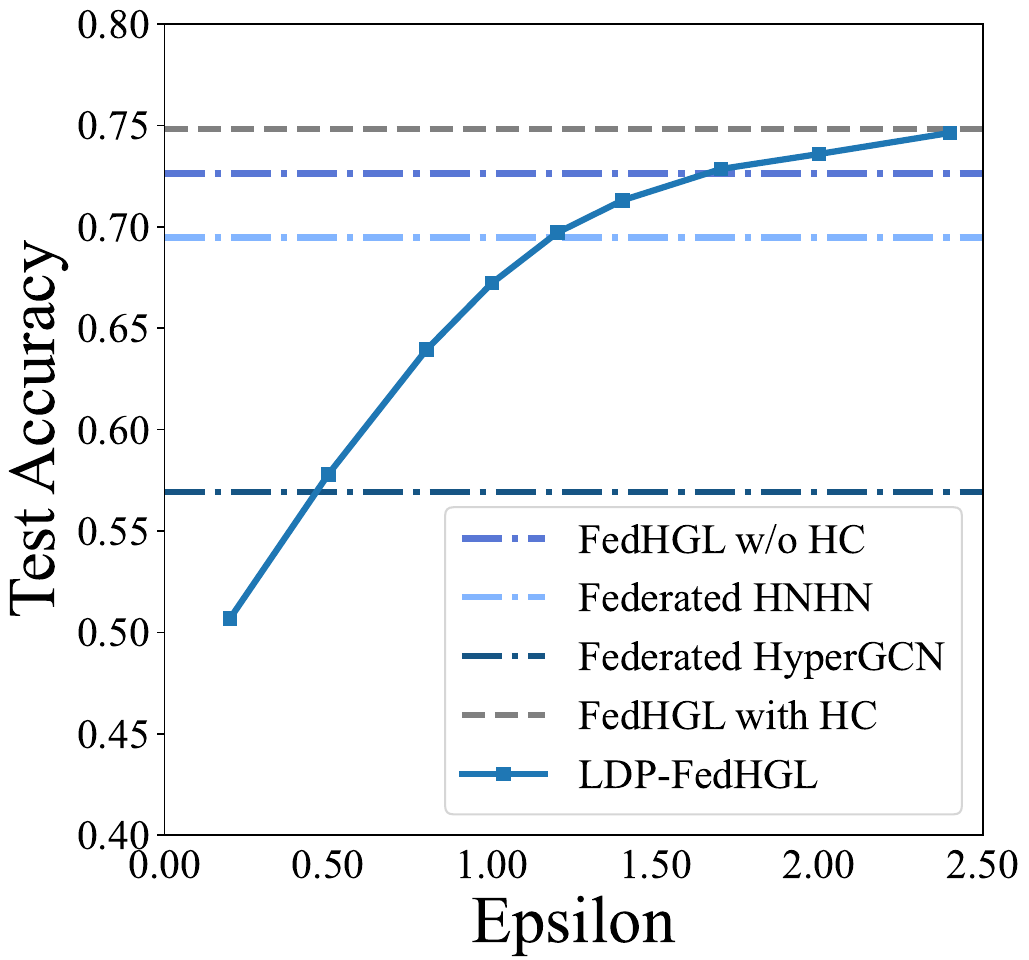}
        \label{fig:DBLPE}
    }

    \subfloat[Cora (Binary).]{%
        \includegraphics[width=1.6in]{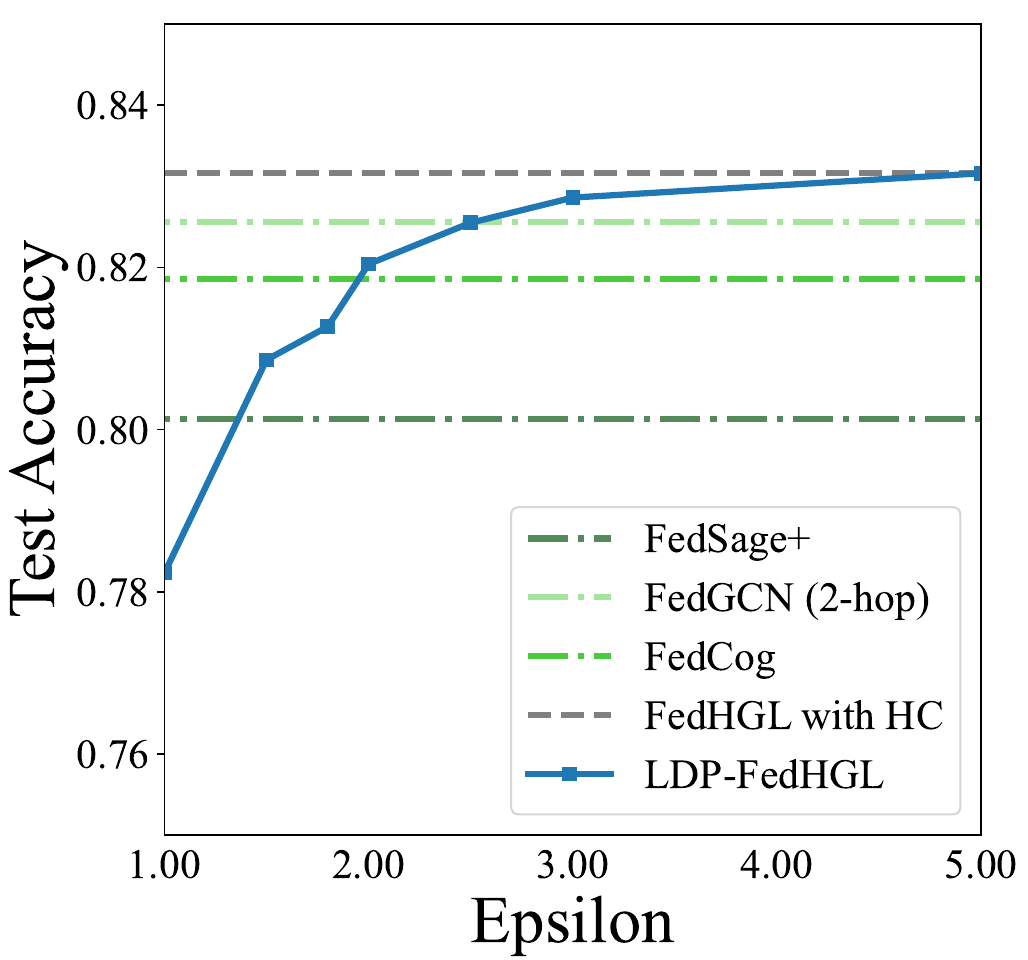}
        \label{fig:CORAE}
    }
    \hfil
    \subfloat[Facebook (Numeric).]{%
        \includegraphics[width=1.6in]{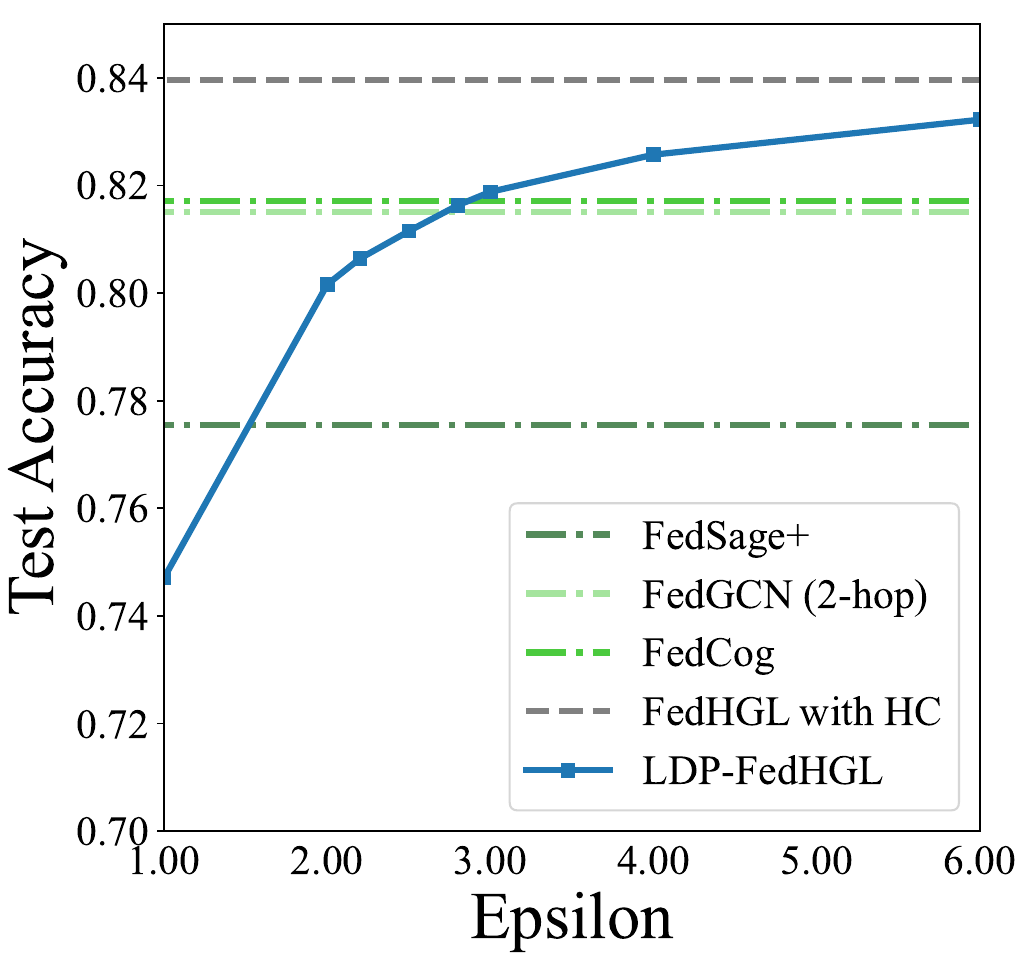}
        \label{fig:FacebookE}
    }

    \caption{Performance of FedHGL with varying privacy budgets.}
    \label{fig:LDP}

\end{figure}

Figure~\ref{fig:LDP} shows the algorithm's performance under different privacy budgets on both hypergraph and simple graph datasets. LDP-FedHGL refers to the variant of FedHGL that incorporates the HC process with a LDP mechanism. Comparison methods that do not use LDP mechanisms are not affected by privacy budgets and are represented as horizontal lines. As the privacy budget increases, the performance of LDP-FedHGL will eventually approach that of FedHGL without the LDP mechanism.

Figure~\ref{fig:CoraCAE} shows the variation in testing accuracy of LDP-FedHGL with the Randomized Response mechanism on the CoraCA dataset as the privacy budget increases. When $ \epsilon > 0.4 $, LDP-FedHGL outperforms the second-best federated HyperGCN, indicating that the algorithm can be adjusted according to the desired range of $\epsilon > 0.4$. Figure~\ref{fig:DBLPE} displays the performance of LDP-FedHGL using the Laplace mechanism on the DBLP dataset, where the algorithm exceeds the second-best FedHGL without HC operation when $ \epsilon > 1.7 $.

Figure~\ref{fig:CORAE} and Figure~\ref{fig:CORAE} demonstrate that LDP-FedHGL outperforms the second-best methods, FedGCN (2-hop) and FedCog, on the Cora and Facebook datasets when $ \epsilon > 2.5 $ and $ \epsilon > 2.8 $, respectively. Note that for federated graph learning methods, when the performance curve of FedHGL with LDP falls below these horizontal lines, it does not necessarily indicate that the method is inferior. This is because these methods either do not address privacy issues or only partially resolve them. From the comparisons on these two datasets and the earlier comparisons on hypergraph datasets, it can be observed that the randomized response mechanism has a lower requirement for privacy budgets.

\subsection{Experimental Infrastructure}

The experiments are conducted on a high-performance computing platform with an Intel Xeon Silver 4310 CPU, an NVIDIA RTX A6000 GPU, and 128GB of RAM. For federated hypergraph mining tasks, PyTorch 2.3.1 was used, ensuring compatibility with CUDA 12.1 for efficient large-scale dataset processing and model training.

\section{Conclusion}

In this work, we present FedHGL, a comprehensive federated hypergraph learning framework designed to address the challenges of cross-client information loss and privacy preservation. Our framework uniquely integrates hypergraph neural networks with federated learning techniques, incorporating a pre-propagation hyperedge completion operation and local differential privacy mechanisms. Leveraging these innovations, FedHGL effectively harnesses high-order information across clients while preserving client privacy. Experiments on real-world datasets demonstrate our algorithm's effectiveness, achieving significant improvements over existing methods.

\section{Acknowledgement}
This work was supported in part by the National Natural Science Foundation of China (Grant no.62302527), in part by the Hunan Provincial Natural Science Foundation (Grant no.2023jj40774), and in part by the High Performance Computing Center of Central South University. 

\bibliographystyle{IEEEtran}
\bibliography{IEEEabrv,references}

\end{document}